\pdfoutput=1

\documentclass[11pt]{article}

\usepackage[final]{acl}

\usepackage{times}
\usepackage{latexsym}

\usepackage[T1]{fontenc}

\usepackage[utf8]{inputenc}

\usepackage{microtype}

\usepackage{inconsolata}

\usepackage{graphicx}
\usepackage{amsmath}

\usepackage{url}
\usepackage{multirow}
\usepackage{amsthm}
\newtheorem{theorem}{Theorem}
\usepackage{amssymb}
\usepackage{amsmath}
\usepackage{xurl}
\usepackage{authblk}

\title{Fighting Spurious Correlations in Text Classification \\via a Causal Learning Perspective}

\author{Yuqing Zhou}
\author{Ziwei Zhu}
\affil{George Mason University\\ Fairfax, VA, USA}
\affil{\{yzhou31, zzhu20\}@gmu.edu}

\begin{document}
\maketitle
\begin{abstract}

In text classification tasks, models often rely on spurious correlations for predictions, incorrectly associating irrelevant features with the target labels. This issue limits the robustness and generalization of models, especially when faced with out-of-distribution data where such spurious correlations no longer hold. To address this challenge, we propose the Causally Calibrated Robust Classifier (CCR), which aims to reduce models' reliance on spurious correlations and improve model robustness. Our approach integrates a causal feature selection method based on counterfactual reasoning, along with an unbiased inverse propensity weighting (IPW) loss function. By focusing on selecting causal features, we ensure that the model relies less on spurious features during prediction. We theoretically justify our approach and empirically show that CCR achieves state-of-the-art performance among methods without group labels, and in some cases, it can compete with the models that utilize group labels. Our code can be found at: \url{https://github.com/yuqing-zhou/Causal-Learning-For-Robust-Classifier}.
\end{abstract}

\section{Introduction}
Despite their success on standard benchmarks, neural networks often struggle to generalize to out-of-distribution (OOD) data. A primary cause of this problem is their tendency to rely on features not causally related to tasks but holding spurious correlations to labels, which can reduce model robustness when the data distribution shifts~\cite{hovy2015tagging, ribeiro2016should, tatman2017gender,buolamwini2018gender,hashimoto2018fairness}. For example, in natural language inference (NLI) tasks, if contradictory sentences in a dataset frequently contain negation words, a model trained on this dataset might predict contradiction simply based on the presence of negation words rather than relying on the true underlying features. When encountering data where such spurious correlations do not hold, the model is likely to make incorrect predictions.

Prior works divide the data into different groups based on combinations of class labels and spurious features. The distribution of these groups is often unbalanced, and the correlation between labels and spurious features in the majority groups leads to spurious correlations. As a result, a model trained on such a dataset inevitably performs worse on the minority groups compared to the majority groups. Existing methods have focused on improving the worst-group accuracy. Some methods tackle this issue by utilizing group labels, which provide information about both the true labels and the spurious features~\cite{sagawa2019distributionally, goel2020model, zhang2020coping, idrissi2022simple}. However, obtaining annotations for group labels is costly and not always feasible, limiting the utility and applicability of these methods. 

Other approaches employ a two-stage process, where group labels are first inferred, followed by an adjustment of the loss function based on this inferred information. Examples include reweighting samples~\cite{liu2021just, qiu2023simple} or incorporating supervised contrastive loss~\cite{zhang2022correct}. However, these methods attempt to calibrate the model learning process by only adjusting the loss function and hope the model's over-reliance on spurious features can be alleviated. The effects are indirect and limited. A more direct and effective approach would involve modifying the model architecture itself to explicitly identify causal features and filter out spurious ones.

Existing works show that models trained using standard Empirical Risk Minimization (ERM) can learn high-quality representations of both spurious and causal features~\cite{kirichenko2022last, izmailov2022feature}. Therefore, our goal is to encourage models to rely more on causal features than spurious ones. We propose a method, Causally Calibrated Robust Classifier (CCR), to identify causal feature representations through counterfactual reasoning. Specifically, we first disentangle the representation vector from the model's last layer and then construct counterfactual representation vectors. We adapt causal inference theory~\cite{pearl2009causality} to calculate the probabilities of necessity and sufficiency for each feature within the representation vector and the entire representation. By retraining the last layer to maximize the probability of necessity and sufficiency of the overall feature representation, the model is guided to assign higher weights to causal features, ultimately relying more on these features. Additionally, we derive a theoretically unbiased loss function based on inverse propensity weighting (IPW), further enhancing the model's robustness to spurious correlations.

Our contributions are summarized as follows:

1. We integrate causal inference theory into feature representation selection by calculating the probabilities of necessity and sufficiency (PNS) for each feature, and propose a method that enhances the weights of causal features in model predictions through counterfactual reasoning by maximizing the PNS for the entire representation vector.

2. We derive a general unbiased loss function with the inverse propensity weighting (IPW) and find that existing methods, which reweight samples to mitigate the effects of spurious correlations, are in fact variants of this unbiased loss function.

3. We propose a Causally Calibrated Robust Classifier (CCR) for text classification tasks that combines causal feature selection with unbiased IPW loss, without the need for group label annotations of training datasets.

4. Comparing our methods CCR with standard ERM, JTT~\cite{liu2021just}, and AFR~\cite{qiu2023simple} on 4 text classification tasks with varying spurious correlations, the proposed method achieves superior performance on most tasks and even outperforms methods that rely on full group labels~\cite{sagawa2019distributionally, kirichenko2022last}.

\section{Method}
In this paper, we focus on the impact of spurious correlations on transformer-based text classifiers and aim to enhance their robustness. While implemented on BERT~\cite{devlin2018bert}, our method can be generalized to other models as well.

Our method, CCR, consists of two stages. In stage 1, we apply standard ERM with a covariance matrix regularization to disentangle the representations in the model's last layer, ensuring each feature representation is independent. In stage 2, we retrain the last layer to encourage the model to rely on causal features by introducing a sample-wise causal constraint based on counterfactual representation vectors. Additionally, we reweight each sample's loss in stage 2 to achieve an unbiased loss, ensuring accurate causal feature selection.

\subsection{Problem Formulation}
In text classification tasks, models are trained to assign a label $y \in \mathcal{Y}$ (such as sentiment, topic, and intent) to a given input text. Transformer-based models, such as BERT, encode the original text input into a fixed-dimensional embedding, regarded as hidden features of the text input. Among these features, some are causally related to the true label, referred to as causal features, while others may be coincidentally correlated with the label, referred to as spurious features. These spurious features do not have a causal connection to the label but can mislead the model. Although models trained by standard ERM can rely on spurious correlation, existing works show that they learn high-quality representations of both causal and spurious features~\cite{kirichenko2022last}.

 We use capital letters ${\bf X}$, ${\bf S}$, and ${\bf Y}$ to denote the variable of causal features, spurious features, and labels, respectively, and ${\bf x}$, ${\bf s}$, and $y$ represent their values. Then a text dataset of N samples can be denoted as $D = \{({\bf x}_0, {\bf s}_0, y_0), ..., ({\bf x}_{N-1}, {\bf s}_{N-1}, y_{N-1})\}$. For simplicity, we assume ${\bf s}_i \in \{{\bf s}^0, {\bf s}^1\}$ where ${\bf s}^0$ and ${\bf s}^1$ represent two distinct spurious features. Then, the dataset $D$ can be divided into groups based on the different combinations of  ${\bf S}$, and ${\bf Y}$. For example, in a binary classification task, the dataset would be divided into four groups: $D_{0,0} = \{({\bf x}_i, {\bf s}^0, y^0)\}$, $D_{0,1} = \{({\bf x}_i, {\bf s}^1, y^0)\}$, $D_{1,0} = \{({\bf x}_i, {\bf s}^0, y^1)\}$, and $D_{1,1} = \{({\bf x}_i, {\bf s}^0, y^1)\}$, where $y^0$ and $y^1$ represent two classes. To avoid ambiguity, in the following text, we will refer to the input text features only as ${\bf x}$ and ${\bf s}$, and use the term "features" exclusively to denote their representation vectors from the model's last layer.

\subsection{Automatic Causal Feature Selection}
In this section, we introduce how to encourage the model to focus on causal features. This is achieved through structural modifications to the model, specifically by incorporating a counterfactual feature selection module, which helps the model prioritize features with true causal significance over those that are merely correlated. 

Existing works show that the final layer of a classifier contains both high-quality representations of spurious and causal features~\cite{kirichenko2022last, izmailov2022feature}. To disentangle these representations, we apply covariance matrix regularization~\cite{cogswell2015reducing}, aiming to maximize the independence of each feature in the final layer. This encourages the separation of spurious and causal features. Next, we conduct automatic causal feature selection on this disentangled feature layer using a counterfactual method. In the following section, we will provide a detailed explanation of this counterfactual approach.

\subsubsection{Probabilistic Causality: Necessity and Sufficiency of Features}
To identify causal features, we will estimate the causality of each feature. To quantify this causality, we introduce the probability of necessity and sufficiency as a metric to estimate the causal effect of features on labels, based on the causation theory of \citet{pearl2009causality}. First, we introduce the definitions of the probability of necessity and sufficiency using the counterfactuals. Let ${\bf E}$ denote the variable of the feature embedding (the last layer of the classifier) and ${\bf e}$ denote its value. Assume the feature embedding vector contains $h$ elements, each representing a feature. Additionally, we define a mask $\textbf{M}$ for the embedding vector, with the same dimensions as ${\bf E}$, where each element $\textbf{M}_i$ takes a value of $1$ or $0$, representing whether a feature is selected ($1$ means the corresponding feature is selected). Thus, the final picked feature representation vector for prediction can be expressed as $\Tilde{\textbf{E}} = \textbf{E} \odot \textbf{M}$, where $\odot$ denotes the element-wise multiplication. 

\textbf{Definition 1.} Probability of Necessity (PN) for a single feature ${\bf E}_j$.
\begin{equation}
\begin{aligned}
    &PN := P(Y_{\textbf{M}_{j} \neq 1, \textbf{M}_{i\neq j} = \textbf{m}_{i\neq j}, (\textbf{X}, \textbf{S}) = (\textbf{x}, \textbf{s})} \neq y | \\ 
     &\quad \textbf{M}_{j} = 1, \textbf{M}_{i\neq j} = \textbf{m}_{i\neq j}, Y = y, (\textbf{X}, \textbf{S}) = (\textbf{x}, \textbf{s}))
\end{aligned}
\label{equ: pn}
\end{equation}
where $\textbf{M}_{j} = 1$ indicates that ${\bf E}_j$ is selected and will contribute to the final prediction, while $\textbf{M}_{j} \neq 1$ indicates that it is not selected.
Equation~\ref{equ: pn} represents the probability of $Y \neq y$ for the same input in the absence of the feature ${\bf E}_j$, given that feature ${\bf E}_j$ exists and $Y = y$ in reality.

\textbf{Definition 2.} Probability of Sufficiency (PS) for a single feature ${\bf E}_j$.
\begin{equation}
\begin{aligned}
     &PS := P(Y_{\textbf{M}_{j} = 1, \textbf{M}_{i\neq j} = \textbf{m}_{i\neq j}, (\textbf{X}, \textbf{S}) = (\textbf{x}, \textbf{s})} = y | \\
     &\quad\textbf{M}_{j} \neq 1, \textbf{M}_{i\neq j} = \textbf{m}_{i\neq j}, Y \neq y, (\textbf{X}, \textbf{S}) = (\textbf{x}, \textbf{s}))
\end{aligned}
\label{equ: ps}
\end{equation}
Equation~\ref{equ: ps} measures the contribution of the feature ${\bf E}_j$ to the prediction of $y$ for the input $(\textbf{x}, \textbf{s})$.

{\bf Definition 3.} Probability of Necessity and Sufficiency (PNS) for a single feature $\textbf{E}_j$.
\begin{equation}
\begin{aligned}
    PNS := P(&Y_{\textbf{M}_{j} \neq 1, \textbf{M}_{i\neq j} = \textbf{m}_{i\neq j}, (\textbf{X}, \textbf{S}) = (\textbf{x}, \textbf{s})} \neq y, \\ 
    &Y_{\textbf{M}_{j} = 1, \textbf{M}_{i\neq j} = \textbf{m}_{i\neq j}, (\textbf{X}, \textbf{S}) = (\textbf{x}, \textbf{s})} = y)
\end{aligned}
\label{equ: pns}
\end{equation}
Equation~\ref{equ: pns} evaluates both the necessity and sufficiency of the feature ${\bf E}_j$ for the label $Y=y$, given the input $(\textbf{x}, \textbf{s})$. If the feature ${\bf E}_j$ is a causal feature in this task, its $PNS$ should have a high value, and We aim to encourage the classifier's final layer to assign a high weight to this feature.

Directly calculating Euqation~\ref{equ: pns} can be difficult. However, by applying and extending Theorem 9.2.10 of \citet{pearl2009causality} to a more general case, we can get a lower bound of PNS.
\begin{theorem} The lower bound of PNS for causal features is given as follows:
\begin{equation}
\begin{aligned}
    &PNS \geq \\
    &\quad max[0,  P(Y_{\textbf{M}_{j} = 1, \textbf{M}_{i\neq j} = \textbf{m}_{i\neq j}, (\textbf{X}, \textbf{S}) = (\textbf{x}, \textbf{s})} = y) \\ 
    &\quad \quad \quad - P(Y_{\textbf{M}_{j} \neq 1, \textbf{M}_{i\neq j} = \textbf{m}_{i\neq j}, (\textbf{X}, \textbf{S}) = (\textbf{x}, \textbf{s})} \neq y)]
\end{aligned}
\label{equ: pns_lower_bound}
\end{equation}
\end{theorem}
We can achieve a higher $PNS$ for the feature ${\bf E}_j$ by maximizing this lower bound $\underline{PNS}$. In practice, $ P(Y_{\textbf{M}_{j} = 1, \textbf{M}_{i\neq j} = \textbf{m}_{i\neq j}, (\textbf{X}, \textbf{S}) = (\textbf{x}, \textbf{s})} = y)$ is estimated from the model's output using the original feature vector, while $P(Y_{\textbf{M}_{j} \neq 1, \textbf{M}_{i\neq j} = \textbf{m}_{i\neq j}, (\textbf{X}, \textbf{S}) = (\textbf{x}, \textbf{s})} \neq y)$ is estimated using the model's output based on the counterfactual feature vector.

\subsubsection{Causal Constraints with PNS}
We introduce a module to generate the counterfactual feature vectors. Let $\textbf{M}$ denote a mask with all values set to $1$, except for one value set to $0$. By $\Tilde{\textbf{E}} = \textbf{E} \odot \textbf{M}$ with $\textbf{M}_j = 0$ for each $j \in \{1, 2, ..., h\}$, we create a counterfactual feature vector, a copy of ${\bf E}$ with the $j$-th element discarded. This process produces $h$ counterfactual feature vectors.

{\bf Definition 4}. Let $PNS_j$ denote the $PNS$ of the $j$-th counterfactual feature vector. We define the total $PNS$ of the feature embedding vector ${\bf E}$ as follows:
\begin{equation}
\begin{aligned}
    &PNS = \prod_{j=1}^{h} PNS_j
\end{aligned}
\label{equ: embeddings_pns}
\end{equation}

We take the mean of the negative log of Equation~\ref{equ: embeddings_pns} as the causal constraint and incorporate it into the loss function, where $PNS_j$ is replaced by its lower bound. In this way, the final layer can be trained to rely more on causal features, i.e., the feature with higher $PNS$ values.

\subsection{Unbiased Loss through IPW}
Accurately selecting causal features requires an unbiased loss. To achieve this goal, we derive a reweighting method using inverse propensity weighting. Previous methods~\cite{liu2021just, qiu2023simple} that use heuristics for weight computation can be regarded as special implementations of our framework.

Assume we have a text classification dataset with $C$ classes $\{y^1, y^2, ..., y^C\}$ and $K$ spurious features $\{{\bf s}^1, ..., {\bf s}^{K}\}$. Then the dataset can be divided into $C * K$ groups, denoted as $D_{j, k} = \{(\textbf{x}_i, \textbf{s}_i, y_i) | y_i = y^j, \textbf{s}_i = \textbf{s}^k)\}$. In an ideal scenario, the size of group $D_{j,k}$ should be same for all $k=1, ..., K,$ within the class $y^j$, i.e., $|D_{j,k}| = |D_{j,t}|$, $\forall k$, $t \in \{1, .., K\}$. In this case, the spurious features ${\bf s}$ are not correlated with the labels $y$. An ideal loss function is as follows:
\vspace{-5pt}
\begin{equation}
\begin{aligned}
    &L_{ideal} = \frac{1}{|D|}\sum_{j=1}^{C}\sum_{k=1}^K L_{D_{j,k}} \\
   & = \frac{1}{|D|}\sum_{j=1}^{C}\sum_{k=1}^K \sum_{i=1}^{|D|} L(y_i, f(\textbf{x}_i, \textbf{s}_i)) \mathbb{I}_{\{y_i =y ^j, \textbf{s}_i=\textbf{s}^k\}},
    \label{equ:ideal_loss}
\end{aligned}
\end{equation}
where $L_{D_{j,k}}$ is the sum of the losses over all samples in the subset $D_{j,k}$,  $|D|$ is the size of the whole dataset $D$, $L(y_i, f(\textbf{x}_i, \textbf{s}_i))$ is the loss on the $i$-th sample, which will be simplified as $L_i$ in the following text. Additionally, $\mathbb{I}$ is an indicator function. $\mathbb{I}_{\{A\}} = 1$ when $A$ is true. Ideally, all subgroups $D_{j,*}$ within the class $y^j$ contribute equally to the total loss. A classifier $f$ learned on this dataset would be robust and make predictions of $y$ solely based on $\textbf{x}$. However, in reality, the group distribution of a dataset $D'$ is unbalanced, i.e., $\exists l$, $|D^{'}_{j,l}| \gg |D^{'}_{j,k}|$, $\forall k\neq l$. In this case, we call $|D^{'}_{j,k}|$ as the majority group when $k=l$, and as the minority group otherwise. A classifier trained on such a dataset would falsely rely on the spurious correlation between $s^l$ and label $y^j$ leading to biased predictions.

We can assume that not all samples in the ideal datasets can be observed and the ratio of samples observed for different groups varies. We use the $\mathcal{O}$ to denote the observation of the whole ideal dataset $D$ and $o_i$ to denote the observation result for $i$-th sample in the ideal dataset, where $o_i = 1$ means the sample is in the reality dataset $D'$, and 0 means the sample is not in $D'$. The probability of $p(o_i=1)$ depends on its group membership, thus leading to an unbalanced dataset $D'$. We use $p_{j,k}(o=1)$ to represent the probability of a sample in $D_{j, k}$ being observed. The majority groups in $D^{'}$ have a much larger $p_{j,k}$ than the minority group. With these notations, the actual loss is given by 
\begin{equation}
    L_{real} = \frac{1}{|D|}\sum_{j=1}^{C}\sum_{k=1}^K \sum_{i=1}^{|D|} L_i \mathbb{I}_{\{y_i =y ^j, \textbf{s}_i=\textbf{s}^k\}} \mathbb{I}_{\{o_i=1\}},
\label{equ: loss_biased}
\end{equation}
In this case, 
\begin{equation}
\begin{aligned}
    &\mathbb{E}[L_{real}] \\
    &= \frac{1}{|D|}\sum_{j=1}^{C}\sum_{k=1}^K \sum_{i=1}^{|D|} L_i \mathbb{I}_{\{y_i =y ^j, \textbf{s}_i=\textbf{s}^k\}} p_{j,k}(o_i=1) \\ 
    &\neq L_{ideal}
    \end{aligned}
\label{equ: expectation_of_biased_loss}
\end{equation}
i.e., $L_{real}$ is biased.

To achieve an unbiased loss on an unbalanced dataset, the key is to balance the weights of the majority and minority groups in the final loss function. This can be achieved by normalizing the losses for each subset and averaging their contributions across the groups. The loss function is defined as
\begin{equation}
\begin{aligned}
    L &= \frac{1}{|D^{'}|} \sum_{j=1}^C \frac{|D_j^{'}|}{K}\sum_{k=1}^K \frac{L_{D^{'}_{j,k}}}{|D^{'}_{j,k}|} \\
    &= \frac{1}{|D^{'}|} \sum_{j=1}^C \frac{1}{K}\sum_{k=1}^K \frac{L_{D^{'}_{j,k}}}{\frac{|D^{'}_{j,k}|}{|D_j^{'}|}}
\end{aligned}
\label{equ:unbiased_loss}
\end{equation}
where $D^{'}_j = \cup_{k=1}^K D^{'}_{j,k}$ and $D^{'} = \cup_{j=1}^C D^{'}_j$. Here, the normalization ensures that within each group $D^{'}_j$, the contributions from subsets $D^{'}_{j,k}$ for $k=1,..., K$ are equal, leading to a more equitable contribution to the overall loss. When the dataset is large enough, we can get
\begin{equation}
\begin{aligned}
    p({\bf S}={\bf s}^k|Y=y^j) = \frac{|D^{'}_{j,k}|}{|D^{'}_j|},
\end{aligned}
\label{equ:propensity}
\end{equation}
which is a propensity for a sample with label $j$ to contain the spurious feature ${\bf s}^k$. This suggests that the Equation~\ref{equ:unbiased_loss} can be expressed in the following form:
\begin{equation}
\begin{aligned}
    L &= \frac{1}{K*|D|}\sum_{j=1}^C \sum_{k=1}^K \sum_{i=1}^{|D|} \frac{L_i \mathbb{I}_{\{y_i =y ^j, \textbf{s}_i=\textbf{s}^k\}} \mathbb{I}_{\{o_i=1\}} }{p({\bf s}^k|Y=y^j)}
\end{aligned}
\label{equ:ideal_IPS}
\end{equation}

As we have no knowledge of group information, so the true $p({\bf s}^k|Y=y^j)$ is unknown. However, if we can find a way to get its estimation $\hat{p}$, then we can get the following loss function:
\begin{equation}
\begin{aligned}
    L_{IPW} &= \frac{1}{|D|}\sum_{j=1}^C \sum_{k=1}^K \sum_{i=1}^{|D|} \frac{L_i \mathbb{I}_{\{y_i =y ^j, \textbf{s}_i=\textbf{s}^k\}} \mathbb{I}_{\{o_i=1\}} }{K*\hat{p}({\bf s}^k|Y=y^j)}
\end{aligned}
\label{equ:real_IPS}
\end{equation}

The expectation of $L_{IPW}$ is given as follows:
\begin{equation}
\begin{aligned}
    &\mathbb{E}[L_{IPW}]\\
    &= \frac{1}{|D|}\sum_{j=1}^C \sum_{k=1}^K \sum_{i=1}^N \frac{L_i \mathbb{I}_{\{y_i =y ^j, \textbf{s}_i=\textbf{s}^k\}} \mathbb{E}[\mathbb{I}_{\{o_i=1\}}] }{K*\hat{p}({\bf s}^k|Y=y^j)} \\
    &= \frac{1}{|D|}\sum_{j=1}^C \sum_{k=1}^K \sum_{i=1}^N \frac{L_i \mathbb{I}_{\{y_i =y ^j, \textbf{s}_i=\textbf{s}^k\}} p(o_i=1)}{K*\hat{p}({\bf s}^k|Y=y^j)}
\end{aligned}
\label{equ:expectation_IPS}
\end{equation}
When $\hat{p}({\bf s}^k|y^j) = \frac{p_{j,k}(o=1)}{K}$, $\mathbb{E}[L_{IPW}] = L_{ideal}$, which is unbiased. It demonstrates that we can get the unbiased loss function through inverse propensity weighting (IPW). The state-of-the-art robust methods JTT~\cite{liu2021just} and AFR~\cite{qiu2023simple} can be considered as special implementations of $L_{IPW}$, which rely on different heuristics for estimating $\hat{p}({\bf s}^k|y^j)$.

Our estimation of $\hat{p}({\bf s}^k|y^j)$ is based on the performance of the first-stage model. For each class, we treat the samples correctly classified by the first-stage model as the major groups (e.g., $(\textbf{s}^1, y^1)$ and $(\textbf{s}^2, y^2)$), while the misclassified samples are taken as minor groups (e.g., $(\textbf{s}^2, y^1)$ and $(\textbf{s}^1, y^2)$). This process divides the dataset into $2C$ groups, and we use each group's proportion to the whole dataset as the estimated $\hat{p}({\bf s}^k|y^j)$.

\subsection{Training Framework}
The overall training process is as follows.

\textbf{Stage 1}. A linear layer is introduced to the pretrained model, with input dimensions $h$ (a hyperparameter, representing the number of features) and output dimensions corresponding to the number of classes. The model is then finetuned using ERM, combining cross-entropy loss with covariance matrix regularization on the feature vectors. This process encourages the model to learn an effective feature extractor, which will be frozen in stage 2, and ensures that the final feature vector consists of $h$ disentangled and independent features.

\textbf{Stage 2}. We freeze all layers of the model except the last one, taking the frozen layers as a fixed feature extractor, while retraining only the last layer in this stage. Before retraining, we evaluate the model's performance on the training set, identifying the correctly and incorrectly classified samples within each class. Based on this, we estimate the data groups and calculate their ratios as the estimated propensities, which are used to assign weights to each group. Then, we finetune the last layer according to the following loss function:
\vspace{-5pt}
\begin{equation}
\begin{aligned}
   L = \frac{1}{|D^{'}|} \sum_{i=1}^{|D^{'}|} \frac{1}{\hat{p}} (&CE(y_i, f({\bf x}_i, {\bf s}_i) - \\ 
   & \lambda * \frac{1}{h}\sum_{j=1}^h\log \underline{PNS}_j)
\end{aligned}
\label{equ: total_loss}
\end{equation}
where the $\lambda$ is the coefficient of causality constraints.

\section{Experiments}
In this section, we investigate the effect of CCR through three research questions. \textbf{RQ1:} What is the performance of CCR compared with other state-of-the-art methods on the text classification benchmarks? \textbf{RQ2:} How is the contribution of each component of CCR to the model's robustness? \textbf{RQ3:} What is the effect of causal feature selection under different hyperparameters?

\subsection{Experiments Setup}
\subsubsection{Datasets}
We evaluate our method, CCR, on two real-world datasets, CivilComments~\cite{borkan2019nuanced} and MultiNLI~\cite{williams2017broad}, following previous works~\cite{liu2021just, qiu2023simple}. Additionally, we use two semi-synthetic datasets, which are real-world datasets manually introduced with spurious correlations: the Yelp dataset~\cite{zhang2015character} with author style shortcuts~\cite{zhou2024navigating} and the Beer dataset~\cite{bao2018deriving} with concept occurrence shortcuts~\cite{zhou2024navigating}. These four datasets cover four types of spurious correlations~\cite{zhou2024navigating}.

\textbf{CivilComments}~\cite{borkan2019nuanced} is a text classification dataset, where the target is to classify a comment as "toxic" or "not toxic". The WILDS benchmark~\cite{koh2021wilds} provides a version of CivilComments that contains spurious correlations between the label "toxic" and some sensitive characteristics such as gender and race, which are used in our experiments. These words describing sensitive characteristics such as gender (male, female) construct a category-word shortcut. 

\textbf{MultiNLI}~\cite{N18-1101} is designed for the task of natural language inference, where each sample contains sentence pairs labeled with one of three categories that describe the relationship between the sentences: entailment, contradiction, or neutral. In this dataset, spurious correlations exist between the label "contradiction" and negation words, which lead to a synonym shortcut.

\textbf{Yelp} review dataset~\cite{zhang2015character} consists of review-rating pairs from Yelp, with ratings ranging from $1$ to $5$. ~\citet{zhou2024navigating} constructed a \textbf{Yelp-Author-Style} dataset based on the Yelp reviews, where each review is rewritten in the distinct styles of different authors (Shakespeare and Hemingway) according to the ratings. We adapt this multiclass classification task into a binary classification task by selecting reviews with ratings of $2$ and $4$, where lower ratings are correlated with Hemingway's style and higher ratings with Shakespeare's style.

\textbf{Beer-Concept-Occurrence} dataset~\cite{zhou2024navigating} consists of review-rating pairs about beer, where each sample contains two parts—one for the palate review (the target) and another representing a spurious feature (comments on either "aroma" or "appearance"). Comments on the appearance of the beer are correlated with lower palate ratings, while comments on the aroma are correlated with higher ratings.

\subsubsection{Model and Baselines}
\paragraph{Model.} Following the settings in AFR~\cite{qiu2023simple}, we use the pretrained model "bert-base-uncased"~\cite{devlin2018bert} for the text classification tasks. The hyperparameter settings are provided in Table~\ref{tab:settings} of Appendix~\ref{app:settings}.

\paragraph{Baselines.} In our experiments, we primarily focus on the following three baselines, which do not require group labels: standard ERM~\cite{liu2021just}, JTT~\cite{liu2021just}, and AFR~\cite{qiu2023simple}. JTT trains two models: the first model is trained via standard ERM, while the second model is trained by upweighting the samples misclassified by the first model. AFR also trains twice but on the same model: it retrains the last layer of a standard ERM-trained model by upweighting the samples on which the model previously performed poorly. 

Additionally, we compare the performance against two baselines that rely on group labels: Group-DRO~\cite{sagawa2019distributionally}, which minimizes the loss of the worst group while applying an additional penalty based on group size to avoid overfitting, and DFR~\cite{kirichenko2022last}, which is similar to AFR but retrains the last layer on a small dataset with group labels.

\subsection{Strong Group Robustness of CCR (RQ1)}
\label{sec: RQ1}
We evaluate CCR on four benchmarks, measuring both overall accuracy and worst-group accuracy (WGA), with the results presented in Table~\ref{tab:overall_results}. Compared to other state-of-the-art robust methods that do not require group labels, CCR achieves the highest WGA across all four datasets, demonstrating strong robustness to spurious correlations. Specifically, CCR achieves the best performance in terms of both mean accuracy over the entire test dataset and WGA on Yelp-Author-Style, improving WGA by $3\%$ and mean accuracy by $0.5\%$, while on Beer-Concept-Occurrence, its mean accuracy is only $0.2\%$ lower than the highest recorded performance. This performance shows that CCR can improve the WGA without sacrificing the overall performance, suggesting its effect on helping the model mitigate the reliance on spurious correlations and utilize causal features.

Furthermore, we compare CCR with two state-of-the-art (SOTA) methods, Group-DRO and DFR, which utilize group labels. As shown in Table~\ref{tab:overall_results1}, CCR achieves comparable performance without the need for group labels in the training data, and even slightly outperforms Group-DRO and DFR on CivilComments, achieving SOTA WGA and a $1\%$ improvement in mean accuracy. It also surpasses DFR in terms of WGA on MultiNLI.

\begin{table*}[ht]
\centering
\begin{tabular}{|c|cc|cc|cc|cc|}
\hline
\multicolumn{1}{|c|}{\textbf{Datasets}} & \multicolumn{2}{c|}{\textbf{CivilComments}}            & \multicolumn{2}{c|}{\textbf{MultiNLI}}                 & \multicolumn{2}{c|}{\textbf{Yelp-Author-Style}}        & \multicolumn{2}{c|}{\textbf{Beer-Concept-Occur}}  \\ \hline
\multicolumn{1}{|c|}{\textbf{Methods}}  & \multicolumn{1}{c|}{Mean}            & WGA             & \multicolumn{1}{c|}{Mean}            & WGA             & \multicolumn{1}{c|}{Mean}            & WGA             & \multicolumn{1}{c|}{Mean}            & WGA             \\ \hline
ERM                                     & \multicolumn{1}{c|}{\textbf{0.9260}} & 0.5740          & \multicolumn{1}{c|}{\textbf{0.8240}} & 0.6790          & \multicolumn{1}{c|}{0.9200}          & 0.8500          & \multicolumn{1}{c|}{0.9538}          & 0.9000          \\ \hline
JTT                                     & \multicolumn{1}{c|}{0.9110}          & 0.6930          & \multicolumn{1}{c|}{0.7860}          & 0.7260          & \multicolumn{1}{c|}{0.9210}          & 0.8583          & \multicolumn{1}{c|}{\textbf{0.9554}} & 0.9000          \\ \hline
AFR                                     & \multicolumn{1}{c|}{0.8980}          & 0.6870          & \multicolumn{1}{c|}{0.8140}          & 0.7340          & \multicolumn{1}{c|}{0.9205}          & 0.8542          & \multicolumn{1}{c|}{0.9506}          & 0.9200          \\ \hline
CCR (ours)                             & \multicolumn{1}{c|}{0.9000}          & \textbf{0.7067} & \multicolumn{1}{c|}{0.8072}          & \textbf{0.7517} & \multicolumn{1}{c|}{\textbf{0.9260}} & \textbf{0.8874} & \multicolumn{1}{c|}{0.9530}          & \textbf{0.9304} \\ \hline
\end{tabular}

\caption{Comparisons of different methods that do not need group labels. We report both the overall mean accuracy and the worst-group accuracy (WGA). The results for ERM, JTT, and AFR on the CivilComments and MultiNLI datasets are sourced from ~\citet{qiu2023simple}. We implement ERM, JTT, and AFR on the Yelp-Author-Style and Beer-Concept-Occurrence datasets. } 
\label{tab:overall_results}
\end{table*}

\begin{table}[h] \footnotesize
\centering
\begin{tabular}{|c|cc|cc|}
\hline
\textbf{Datasets} & \multicolumn{2}{c|}{\textbf{CivilComments}}          & \multicolumn{2}{c|}{\textbf{MultiNLI}}               \\ \hline
\textbf{Methods}  & \multicolumn{1}{c|}{Mean}           & WGA            & \multicolumn{1}{c|}{Mean}           & WGA            \\ \hline
Group-DRO         & \multicolumn{1}{c|}{0.889}          & 0.699          & \multicolumn{1}{c|}{0.814}          & \textbf{0.777 }         \\ \hline
DFR               & \multicolumn{1}{c|}{0.872}          & 0.701          & \multicolumn{1}{c|}{\textbf{0.821}} & 0.747          \\ \hline
CCR(ours)        & \multicolumn{1}{c|}{\textbf{0.900}} & \textbf{0.707} & \multicolumn{1}{c|}{0.807}          & 0.752 \\ \hline
\end{tabular}

\caption{Comparing CCR with the methods need group labels. The results for Group-DRO and DFR are sourced from ~\citet{qiu2023simple}.}
\label{tab:overall_results1}
\end{table}

\subsection{Ablation Study (RQ2)}
After showing the effect of CCR on improving model robustness, we aim to understand how each component of CCR contributes to the overall performance. Using the Yelp-Author-Style dataset, we investigate the impact of the following operations: feature representation disentanglement, causal feature constraints (CFC) for causal feature selection (CFS), and loss debiasing with inverse propensity weighting (IPW). The results are shown in Table~\ref{tab:ablation} and we have the following observations from it:

First, representation disentanglement, causal feature selection (CFS), and loss debiasing with IPW all contribute to improving the model's robustness compared to standard ERM, although the improvements are subtle. CFS alone increases WGA by only $0.8\%$, but when combined with feature disentanglement, it achieves a $2\%$ improvement in WGA. This suggests that disentangling spurious and causal features as much as possible is essential. If these features are entangled, it may diminish the evaluation of each feature's necessity and sufficiency to labels, limiting the effectiveness of CFS.

Second, CFS and IPW all get more than $1\%$ improvement in WGA by combining with disentanglement. Meanwhile, disentanglement alone can improve the WGA of ERM by $1.6\%$, outperforming CFS and IPW. These results suggest that disentanglement may transform the input features into a more effective representation space, which CFS and IPW can leverage to enhance performance.

Third, while both CFS and IPW individually enhance the model's robustness, their combination is the most effective. After disentanglement, the joint effect of CFS and IPW improves both the mean accuracy and WGA of the model, even increasing WGA by approximately $4\%$.

Additionally, we explore three different approaches for propensity estimation in IPW: one proposed by us and two others derived from AFR and JTT's methods for calculating weights. Our method involves counting the samples correctly and incorrectly classified by the first-stage model within each class, then using the inverse of their ratio as the weights, normalized across the entire training dataset. All three methods show similar performance.

\begin{table}[h]\footnotesize
\centering
\begin{tabular}{|l|c|c|}
\hline
\multicolumn{1}{|c|}{\textbf{Methods}} & Mean   & WGA    \\ \hline
ERM                                    & 0.9200 & 0.8500 \\ \hline
disentangle                            & 0.9235 & 0.8658 \\ \hline
CFS                                    & 0.9210 & 0.8583 \\ \hline
IPW                                    & 0.9210 & 0.8571 \\ \hline
disentangle + CFS                      & 0.9235 & 0.8701 \\ \hline
disentangle + IPW                      & 0.9245 & 0.8701 \\ \hline
disentangle + CFS + IPW (ours)        & 0.9260 & 0.8874 \\ \hline
disentangle + CFS + IPW (AFR)         & 0.9265 & 0.8916 \\ \hline
disentangle + CFS + IPW (JTT)         & 0.9250 & 0.8831 \\ \hline
\end{tabular}
\caption{Comparisons of the effect of each component in CCR on the Yelp-Author-Style dataset. For simplicity, we omit ERM from each row. CFS refers to the module for causal feature selection. IPW (ours) represents our proposed method for obtaining weights, while IPW (AFR) and IPW (JTT) use the weight calculation algorithms from AFR~\cite{qiu2023simple} and JTT~\cite{liu2021just}, respectively.}
\label{tab:ablation}
\end{table}

\subsection{Hyperparameter Study (RQ3)}
\begin{figure}[]
    \centering
    \includegraphics[width=0.4\textwidth]{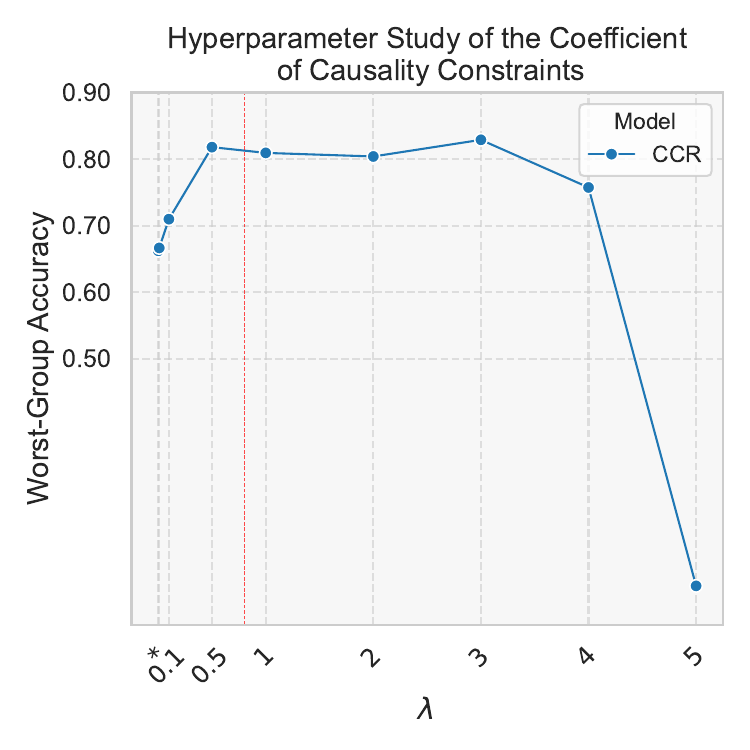}
    \caption{Study on the effect of different coefficients for causality constraints in CFS.($*$ = 0.001)}
    \label{fig:hyper_reg_causal}
\end{figure}

In this section, we investigate the effect of the causality constraint coefficient $\lambda$ in Equation~\ref{equ: total_loss} on model performance. Specifically, we aim to investigate how much emphasis should be placed on the PNS of the feature representation in the final loss function for optimal performance. The experiment results on the Yelp-Author-Style dataset are shown in Figure~\ref{fig:hyper_reg_causal}. From Figure~\ref{fig:hyper_reg_causal}, we can see that as $\lambda$ increases from $0.001$ to $0.5$, WGA increases gradually. However, adding more PNS constraints does not necessarily lead to better performance. As $\lambda$ increases from $0.5$ to $3$, WGA fluctuates slightly, reaching its peak at $\lambda=3$, but gradually decreases as $\lambda$ increases further.

\subsection{Analyzing Model Behavior Through Explainability}
In Section~\ref{sec: RQ1}, we demonstrated that CCR improves model robustness against spurious correlations. In this section, we further explore model behavior through explainability techniques. We use the SHAP~\cite{lundberg2017unified} analysis tool to examine how classifiers rely on specific tokens for their predictions. 

We use the Beer-Concept-Occurrence dataset as an example, where descriptions of beer appearance are correlated with low palate ratings, while descriptions of beer aroma are associated with high palate ratings. We sampled 200 test instances and computed the average SHAP absolute values for the spurious features (i.e., comments on beer appearance and aroma). Table~\ref{tab:shap} shows the SHAP analysis results for each method. The SHAP absolute value quantifies a token's influence on the model's prediction for the corresponding label. For the spurious features, SHAP values closer to zero indicate less reliance on these features by the model. Although the task is relatively straightforward and all methods yield low SHAP values, we observe that CCR’s SHAP values for spurious features are significantly lower than those of other methods—approximately an order of magnitude lower—demonstrating strong robustness and effectively reducing reliance on spurious features in predictions. Besides, we provide an example of SHAP analysis in Appendix~\ref{app:Explainability}.

\begin{table}[]
\centering
\begin{tabular}{|c|cc|}
\hline
\multirow{2}{*}{\textbf{Methods}} & \multicolumn{2}{c|}{\textbf{Label}}      \\ \cline{2-3} 
                                  & \multicolumn{1}{c|}{0}        & 1        \\ \hline
ERM                               & \multicolumn{1}{c|}{1.63E-3} & 1.54E-3 \\ \hline
JTT                               & \multicolumn{1}{c|}{3.10E-3} & 2.54E-3 \\ \hline
AFR                               & \multicolumn{1}{c|}{1.47E-3} & 1.47E-3 \\ \hline
CCR (Ours)                              & \multicolumn{1}{c|}{\textbf{1.85E-4}} & \textbf{1.85E-4} \\ \hline
\end{tabular}
\caption{SHAP values different methods}
\label{tab:shap}
\end{table}

\section{Related Work}
Spurious correlations challenge the robustness of machine learning models trained via ERM, especially when there are shifts between the training and test data distribution~\cite{hovy2015tagging, ribeiro2016should, tatman2017gender,buolamwini2018gender,hashimoto2018fairness}. 

 ~\citet{sagawa2019distributionally} addresses this problem by training a classifier via a regularized group distributionally robust optimization (Group-DRO) method, which minimizes the loss of the worst-case group. This approach is to train the entire model. In contrast, some methods propose that simply retraining the last layer of an ERM-trained model can effectively improve robustness. For example, DFR~\cite{kirichenko2022last} retrain the last layer of the model trained via ERM on a small dataset where the spurious correlations break. However, both \citet{sagawa2019distributionally} and ~\citet{kirichenko2022last} require prior knowledge of spurious correlations to define groups in the training data, making it costly and limiting its ability to generalize to unseen spurious attributes. Some methods upweight samples during retraining where the initially trained model performs poorly~\cite{liu2021just, qiu2023simple}. Another approach uses contrastive learning to improve robustness by aligning same-class representations and separating misclassified negatives, allowing the model to learn similar representations of same-class samples and ignore spurious attributes~\cite{zhang2022correct}. These methods do not require prior knowledge of group labels. All these approaches focus on reducing spurious correlations by modifying the loss function, while our method tackles the underlying issue by directly identifying causal features using a counterfactual feature generation module. 

Our method for identifying causal features is based on the theorem of probability of necessity and sufficiency from causality theory~\cite{pearl2009causality}. A related work also applies this theory~\cite{pmlr-v202-zhang23ap}, but their goal is to identify minimal tokens in input text that lead to correct predictions. In contrast, our approach operates on feature representations within the model to identify causal features, aiming to build a robust classifier that performs well across varying data distributions.

\section{Conclusion}
In this paper, we address the problem of models relying on spurious correlations in their predictions, particularly in text classification tasks. To mitigate this issue, we propose the Causally Calibrated Robust Classifier (CCR). Specifically, we introduce a causal feature selection model based on counterfactual reasoning, combined with an unbiased inverse propensity weighting (IPW) loss function. We provide theoretical justification for our method and empirically demonstrate that it achieves state-of-the-art performance among approaches that do not require group labels, and in some cases, it can even compete with models that utilize group labels.

\section{Limitations}
Apart from the contributions mentioned in the paper, there are still some limitations of this work. 

First, the loss debiasing relies on the quality of the estimation of the propensity $\hat{p}$. If $\hat{p}$ is not accurately estimated, the loss function may still exhibit bias. Improving the estimation method could lead to enhanced performance.

Second, our experiments showed that the optimal feature embedding size varies across datasets, with different sizes yielding the best results. Identifying the appropriate embedding size is important for effectively separating irrelevant features from causal features.

Additionally, we did not explore alternative disentanglement methods, which may improve feature representation. Better disentanglement could lead to more effective causal feature selection and ultimately enhance model performance.

\section*{Acknowledgements}
This work was partially supported by resources provided by the Office of Research Computing at George Mason University (URL: https://orc.gmu.edu) and funded in part by grants from the Commonwealth Cyber Initiative.

\section*{Ethics Statement}
Our research complies with ethical standards, utilizing publicly available datasets. All contents in the datasets do NOT represent the authors' views.

\nocite{Ando2005,andrew2007scalable,rasooli-tetrault-2015}


\bibliography{custom}

\appendix

\section{Appendix}
\label{sec:appendix}

\subsection{Experiment Setup}
\label{app:settings}
The reported performance was achieved using the settings outlined in Table~\ref{tab:settings}. All experiments were conducted once, with a fixed random seed to ensure reproducibility. The pre-trained model, 'bert-base-uncased,' consists of approximately 110M parameters. The statistics for each dataset are provided in Table~\ref{tab:datasets_info}. Each experiment can be completed within 24 hours on a single A100 GPU. 

\begin{table*}[]\footnotesize
\centering
\begin{tabular}{|l|c|c|c|c|}
\hline
\multicolumn{1}{|c|}{\textbf{Datasets}}     & CivilComments & MultiNLI & Yelp-Author-Style & Beer-Concept-Occurrence \\ \hline
\textbf{Coefficient for disentanglement}    & 0.5           & 0.5      & 0.5               & 0.5                     \\ \hline
\textbf{Coefficient for causal constraints} & 0.1           & 2        & 3                 & 2                       \\ \hline
\textbf{Learning rate}                      & 2.00E-02      & 2.00E-01 & 3.00E-03          & 3.00E-02                \\ \hline
\textbf{Weight decay}                       & 0             & 0        & 1.00E-04          & 1.00E-04                \\ \hline
\textbf{Batch size}                         & 32            & 32       & 32                & 32                      \\ \hline
\textbf{Size of Feature Embedding}          & 128           & 128      & 128               & 150                     \\ \hline
\textbf{Epoch}                              & 15            & 10       & 30                & 30                      \\ \hline
\textbf{Seed}                               & 42            & 42       & 42                & 42                      \\ \hline
\end{tabular}
\caption{The hyperparameter settings for each dataset.}
\label{tab:settings}
\end{table*}

\subsection{Explainability Analysis}
\label{app:Explainability}
\begin{figure}[t!]
    \centering
    \includegraphics[width=0.45\textwidth]{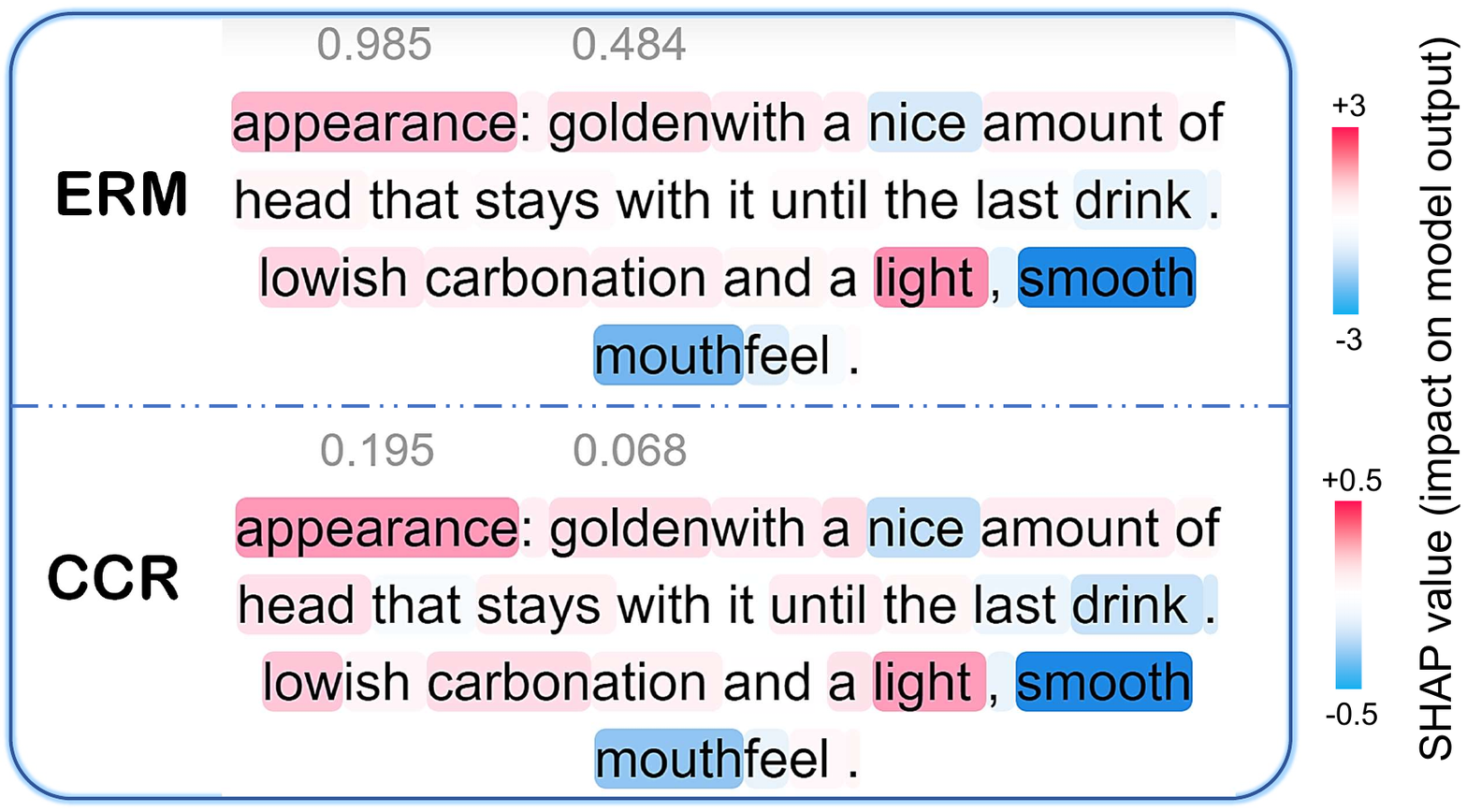}
    \vspace{-10pt}
    \caption{SHAP Analysis for BERT with ERM and CCR}
    \label{fig:shap}
    \vspace{-10pt}
\end{figure}

Figure~\ref{fig:shap} presents a review of beer palate labeled as $0$ (corresponding to a $0.6$ palate rating), along with the behaviors of models trained using standard ERM and CCR in predicting this label. In the ERM model, there is a strong reliance on the words that describe "appearance" (e.g., "appearance" and "golden"). These words show high SHAP values (highlighted in red), indicating that the model attributes significant importance to these terms when predicting a label of $0$. In contrast, the CCR model exhibits much lower SHAP values for the same appearance-related terms, indicating that CCR has successfully reduced the model's reliance on the spurious correlation between appearance-related features and label $0$. The SHAP values in the CCR model are closer to neutral (around zero), showing that the model is less biased toward irrelevant features and more robust to spurious correlations compared to the standard ERM model.

\begin{table*}[]\footnotesize
\centering
\begin{tabular}{|c|c|c|l|c|}
\hline
\multicolumn{1}{|l|}{\textbf{}}              & \textbf{Group IDs} & \multicolumn{1}{l|}{\textbf{Class Label}} & \multicolumn{1}{c|}{\textbf{Group (spurious feature, class)}} & \textbf{\# Training Data} \\ \hline
\multirow{4}{*}{\textbf{CivilComments}}      & 0                  & 0                                         & (no identities, non-toxic)                                    & 148486                    \\ \cline{2-5} 
                                             & 1                  & 0                                         & (has identities, non-toxic)                                   & 90337                     \\ \cline{2-5} 
                                             & 2                  & 1                                         & (no identities, toxic)                                        & 12731                     \\ \cline{2-5} 
                                             & 3                  & 1                                         & (hasidentities, toxic)                                        & 17784                     \\ \hline
\multirow{6}{*}{\textbf{MultiNLI}}           & 0                  & 0                                         & (no negations, contradiction)                                 & 57498                     \\ \cline{2-5} 
                                             & 1                  & 0                                         & (has negations, contradiction)                                & 11158                     \\ \cline{2-5} 
                                             & 2                  & 1                                         & (no negations, entailment)                                    & 67376                     \\ \cline{2-5} 
                                             & 3                  & 1                                         & (has negations, entailment)                                   & 1521                      \\ \cline{2-5} 
                                             & 4                  & 2                                         & (no negations, neutral)                                       & 66630                     \\ \cline{2-5} 
                                             & 5                  & 2                                         & (has negations, neutral)                                      & 1992                      \\ \hline
\multirow{4}{*}{\textbf{Yelp-Author-Style}}  & 0                  & 0                                         & (Hemingway's Style, Rating 2)                                 & 684                       \\ \cline{2-5} 
                                             & 1                  & 0                                         & (Shakespeare's Style, Rating 2)                               & 214                       \\ \cline{2-5} 
                                             & 2                  & 1                                         & (Hemingway's Style, Rating 4)                                 & 252                       \\ \cline{2-5} 
                                             & 3                  & 1                                         & (Shakespeare's Style, Rating 4)                               & 660                       \\ \hline
\multirow{4}{*}{\textbf{Beer-Concept-Style}} & 0                  & 0                                         & (Comments on Appearance , Rating 0.6)                          & 477                       \\ \cline{2-5} 
                                             & 1                  & 0                                         & (Comments on Aroma, Rating 0.6)                               & 261                       \\ \cline{2-5} 
                                             & 2                  & 1                                         & (Comments on Appearance,  Rating   1.0)                        & 65                        \\ \cline{2-5} 
                                             & 3                  & 1                                         & (Comments on Aroma, Rating 1.0)                               & 432                       \\ \hline
\end{tabular}
\caption{The statistics of each dataset. The statistics for CivilComments and MultiNLI are sourced from~\cite{qiu2023simple}.}
\label{tab:datasets_info}
\end{table*}

\end{document}